\documentclass[lettersize,journal]{IEEEtran}
\usepackage{amsmath,amsfonts}
\usepackage{algorithmic}
\usepackage{algorithm}
\usepackage{array}
\usepackage[caption=false,font=normalsize,labelfont=sf,textfont=sf]{subfig}
\usepackage{textcomp}
\usepackage{booktabs}
\usepackage{stfloats}
\usepackage{url}
\usepackage{verbatim}
\usepackage{graphicx}
\usepackage{cite}
\usepackage{multicol}
\usepackage{multirow}
\usepackage{pifont}
\begin{document}

\title{Observation-Guided Meteorological Field Downscaling at Station Scale: A Benchmark and a New Method}

\author{Zili Liu, Hao Chen$^{\dag}$~\IEEEmembership{Member,~IEEE}, Lei Bai, Wenyuan Li, Keyan Chen, Zhengyi Wang, Wanli Ouyang, Zhengxia Zou and Zhenwei Shi$^{\dag}$~\IEEEmembership{Senior Member,~IEEE}
\thanks{The work was supported by the National Natural Science Foundation of China under Grants 62125102, the National Key Research and Development Program of China (Grant No. 2022ZD0160401), the Beijing Natural Science Foundation under Grant JL23005, and the Fundamental Research Funds for the Central Universities, the National Key Research and Development Program of China(Grant No.2022ZD0160101). (Corresponding author: Zhenwei Shi (e-mail:shizhenwei@buaa.edu.cn))}
\thanks{Zili Liu, Keyan Chen and Zhenwei Shi are with the Image
Processing Center, School of Astronautics, with the Beijing Key Laboratory
of Digital Media, and with the State Key Laboratory of Virtual Reality
Technology and Systems, Beihang University, Beijing 100191, China, and
also with the Shanghai Artificial Intelligence Laboratory, Shanghai 200232,
China.

Hao Chen, Lei Bai, and Wanli Ouyang are with Shanghai Artificial Intelligence Laboratory, Shanghai 200232, China.

Wenyuan Li is with the Department of Geography, University of Hong Kong, Hong Kong, China.

Zhengyi Wang is with the School of Oceanography, Shanghai Jiao Tong University, Shanghai 200030, China, and also with Shanghai Artificial Intelligence Laboratory, Shanghai 200232, China.

Zhengxia Zou is with the Department of Guidance, Navigation and Control,
School of Astronautics, Beihang University, Beijing 100191, China, and also
with Shanghai Artificial Intelligence Laboratory, Shanghai 200232, China.}}

\markboth{Journal of \LaTeX\ Class Files,~Vol.~14, No.~8, August~2021}%
{Shell \MakeLowercase{\textit{et al.}}: A Sample Article Using IEEEtran.cls for IEEE Journals}


\maketitle

\begin{abstract}
Downscaling (DS) of meteorological variables involves obtaining high-resolution states from low-resolution meteorological fields and is an important task in weather forecasting. Previous methods based on deep learning treat downscaling as a super-resolution task in computer vision and utilize high-resolution gridded meteorological fields as supervision to improve resolution at specific grid scales. However, this approach has struggled to align with the continuous distribution characteristics of meteorological fields, leading to an inherent systematic bias between the downscaled results and the actual observations at meteorological stations. In this paper, we extend meteorological downscaling to arbitrary scattered station scales, establish a brand new benchmark and dataset, and retrieve meteorological states at any given station location from a coarse-resolution meteorological field. Inspired by data assimilation techniques, we integrate observational data into the downscaling process, providing multi-scale observational priors. Building on this foundation, we propose a new downscaling model based on hypernetwork architecture, namely \emph{HyperDS}, which efficiently integrates different observational information into the model training, achieving continuous scale modeling of the meteorological field. Through extensive experiments, our proposed method outperforms other specially designed baseline models on multiple surface variables. Notably, the mean squared error (MSE) for wind speed and surface pressure improved by 67\% and 19.5\% compared to other methods. We will release the dataset and code subsequently.
\end{abstract}


\section{Introduction}
\IEEEPARstart{I}{n} recent years, the application of artificial intelligence technologies such as deep learning in weather forecasting has garnered significant attention \cite{ren2021deep, mukkavilli2023ai,chen2023foundation}. These efforts utilize large-scale gridded historical meteorological field data, combined with advanced models from computer vision, and have demonstrated powerful performance in many forecasting tasks \cite{bi2023accurate,lam2023learning,chen2023fengwu,chen2023fuxi}, even surpassing the long-developed numerical weather prediction systems \cite{bauer2015quiet}. Nevertheless, the reliance on methodologies from the field of computer vision has led to the acquisition of image-like meteorological field data in the form of fixed and relatively coarse resolution. This approach is at odds with the intrinsically multi-scale nature of meteorological variables. Consequently, to obtain meteorological variables at varying scales and resolutions, \emph{downscaling} has become an indispensable post-processing task within operational forecasting \cite{chen2023foundation, ren2023superbench}.

The objective of downscaling in weather forecasting is typically to map coarse-resolution global-scale meteorological fields to high-resolution regional-scale fields \cite{sun2024deep}. This setup appears to be highly analogous to the task of image super-resolution in computer vision \cite{wang2020deep}. The traditional dynamic downscaling methods \cite{xu2019dynamical} are akin to numerical weather prediction techniques, involving the numerical solution of atmospheric differential equations at a regional scale. This process is highly computationally demanding, particularly when the grid resolution is very high. As a result, machine learning-based downscaling methods, primarily those involving deep learning, have recently received increased attention as a parallel approach \cite{sun2024deep, chen2023foundation,rampal2022high,harris2022generative,kumar2023modern, wang2021deep, sha2020deep,sha2020deep_1,liu2023statistical,hohlein2020comparative,liu2020climate,vandal2017deepsd,vandal2019intercomparison,van2023deep,zhonginvestigating,mardani2023generative,leinonen2020stochastic,harris2022generative,annau2023algorithmic,ren2023superbench,li2023robust}. 
Most previous deep learning-based downscaling works directly employ models and methods from image super-resolution tasks due to the similarity between the two tasks. Meteorological field data is treated as an image to achieve super-resolution at a fixed resolution and fixed upscaling factor, which is direct and efficient. However, this direct application of existing methods also means that the models lack specialized design and flexibility tailored to the unique characteristics of meteorological data.

Unlike natural images that are directly captured through camera sensors, gridded meteorological field data are obtained by fusing and assimilating multi-source, multi-scale, and multi-modal observational and forecast data, typically referred to as \emph{analysis} data or \emph{reanalysis} data. The observational information employed generally includes satellite remote sensing images, ground observation stations, radiosondes, and so on. A specific meteorological variable at a particular pixel can be considered as the average of all observed and predicted values within that pixel area. For instance, the widely used ERA-5 reanalysis data \cite{hersbach2020era5} have a temporal resolution of 1 hour and a spatial resolution of $0.25^{\circ}$. The state value of any given pixel can be regarded as the average of all observations and forecast results within the $0.25^{\circ} \times 0.25^{\circ} $ grid over 1 hour. The same applies to forecast fields derived from analysis and reanalysis data. As a result, for meteorological fields with fixed resolution, although each grid can be considered as the average of all observations, many sub-grid observations cannot be effectively represented. 

However, in practical applications, we aim to obtain the meteorological state at a specific precise location through the given gridded meteorological field, rather than merely obtaining the high-resolution gridded meteorological fields. The absence of sub-grid information results in a significant deviation between the state values of the meteorological field and the scattered stations \cite{wu2023interpretable}. Therefore, for the downscaling of meteorological variables, it is crucial to determine how to increase resolution while accurately recovering information at the sub-grid scale. A straightforward method of recovering sub-grid information is to use high-resolution real-time observational data with multi-scale resolution to guide the downscaling task. Most of the existing downscaling frameworks based on deep learning are inspired by super-resolution tasks. They achieve downscaling solely by learning the mapping from low-resolution to high-resolution images, which does not allow for the integration of multi-scale observations into the model's training and inference processes. With its fixed super-resolution factor, the resulting downscaled output is also gridded and does not provide a continuous representation of the meteorological field. This makes it challenging to generalize well to scattered stations. Additionally, to the best of our knowledge, there currently exists no unified benchmark and dataset for the downscaling of meteorological fields available to researchers in the field. Thus, establishing a reasonable downscaling task specifically tailored to the meteorological field is of critical importance. 

\begin{figure}
\centering
\includegraphics[width=\linewidth]{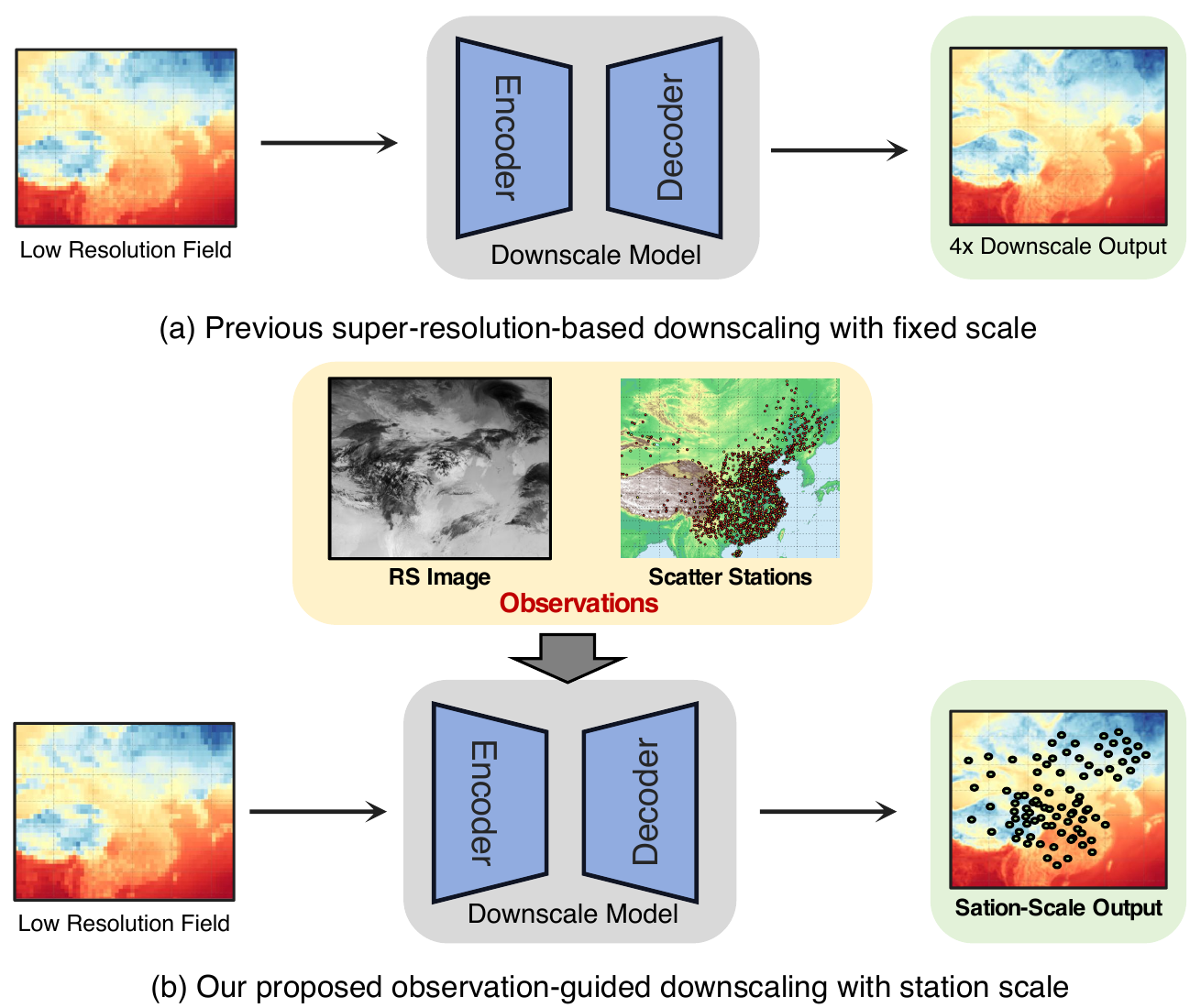}
\caption{The difference between the previous SR (super-resolution)-based downscaling pipeline with fixed grid-level scale \cite{sun2024deep} (a), and the proposed observation-guided downscaling pipeline with arbitrary scatter station-level scale.}
\label{fig:overview}
\end{figure}

To address the aforementioned issues, our paper first constructs a benchmark for downscaling meteorological fields, guided by multi-source, multi-scale, and multi-modal observational data. Our goal is to downscale low-resolution meteorological fields to the scale of arbitrary scatter points. Specifically, the paper selects ERA5 reanalysis data \cite{hersbach2020era5} as the meteorological field data to be downscaled. For observational data, we utilize remote sensing images from the new generation geostationary meteorological satellite Himawari-8 (H8) \cite{bessho2016introduction} at L1-level as high-resolution gridded-scale indirect observational data. Additionally, we employ meteorological observation station data obtained from the Weather2K dataset \cite{zhu2023weather2k} as direct observational data at the scatter station scale. This task setting is crucial for the downscaling of meteorological variables because it allows for the utilization of multi-scale observational information, and it can yield downscaling results that are adaptable to multiple scales. The difference between the previous downscaling task and the proposed benchmark is illustrated in Fig. \ref{fig:overview}.

In response to the established benchmark, inspired by the ability of implicit neural representation methods \cite{xie2022neural} in computer vision to continuously model two-dimensional and three-dimensional data, and combined with a data-conditioned hypernetworks structure \cite{chauhan2023brief}, we propose a novel model for the continuous resolution downscaling of meteorological fields, named \emph{HyperDS}. \emph{HyperDS} uses H8 observations as the model's data-conditioned input and takes Weather2K station data as supervision at the scatter station scale. The overall architecture of \emph{HyperDS} can be divided into a \emph{dual-branch hypernetwork} and a \emph{target network}. The former consists of two encoders based on convolutional neural networks that are used to extract semantic features from the low-resolution meteorological field and H8 data, respectively. Following these, an \emph{implicit retrieval model} employs a cross-attention mechanism to implicitly learn the retrieval process from satellite imagery to meteorological fields, thereby efficiently integrating H8 data into the downscaling process. This results in the generation of high-level feature vectors that contain fused information. The target network is based on a multilayer perceptron (MLP), whose weights are generated from the fused features output by the hypernetwork. It achieves continuous-resolution downscaling at arbitrary locations by inputting the coordinate values of the target location and obtaining the corresponding meteorological state variables. We have also devised a training technique utilizing sub-grid sampling, and in combination with supervisory data from observational stations, it effectively reconstructs accurate state values for meteorological variables at the sub-grid level.

We established several baseline methods and compared them with the proposed method under the condition of identical input and supervision data. \emph{HyperDS} shows superior downscaling performance at the scatter station scale. Additionally, through more extensive analysis and ablation experiments, we also verified the importance of incorporating observational data for the task of downscaling with multi-scale generalization. We hope that more researchers in the field will engage in further studies on this new benchmark, aiming to achieve more efficient continuous-resolution modeling of meteorological fields and more effective integration of observational data. To summarize, the main innovative contributions of this paper include the following three points:
\begin{itemize}
    \item Taking into account the characteristics of the meteorological variables, we have redesigned a new downscaling benchmark that integrates multiple observations into the downscaling process, enabling arbitrary scatter stations downscaling and continuous meteorological field modeling.
    \item Based on this new benchmark, we propose a novel model structured around a data-conditioned hypernetworks architecture, namely \emph{HyperDS}, which achieves scatter station-scale downscaling of meteorological fields.
    \item Through the design of fair baseline models and extensive experiments, we have validated the effectiveness of the proposed new model, which significantly outperforms comparative methods at the scale of scattered stations.
\end{itemize}

\section{Related Work}
In this section, we will briefly introduce and review the work related to the proposed benchmark and model in this paper.
\subsection{Meteorological Field Downscaling}
The objective of downscaling meteorological fields is to obtain accurate weather forecast results with fine granularity and high resolution as required \cite{sun2024deep, chen2023foundation}. Typically, meteorological forecast data generated by operational global forecasting systems are produced on a relatively coarse-resolution grid. Currently, the highest resolution global forecast and reanalysis data are provided by the European Centre for Medium-Range Weather Forecasts (ECMWF), with their operational Integrated Forecasting System (IFS) and ERA-5 analysis \cite{hersbach2020era5} with a spatial resolution of $0.25^{\circ}$ and temporal resolution of 1 hour. To obtain higher-resolution regional-scale weather states, there have been numerous downscaling efforts in the past. Here, we focus our research on methods based on deep learning.

Due to the similarity between the task of downscaling and the task of super-resolution in the field of computer vision, the vast majority of previous work has been inspired by related efforts in the field of super-resolution. One of the most mainstream approaches is the use of super-resolution networks, such as UNet \cite{siddique2021u}, with an encoder-decoder structure based on CNN \cite{rampal2022high, wang2021deep, sha2020deep, sha2020deep_1,liu2023statistical,hohlein2020comparative,liu2020climate,vandal2017deepsd,vandal2019intercomparison,van2023deep} and Transformer \cite{zhonginvestigating}. Furthermore, due to the successful application of generative modeling techniques in the field of super-resolution, many previous studies have also applied Generative Adversarial Networks (GANs) and Diffusion models to the task of downscaling meteorological fields, to obtain results with richer texture information \cite{mardani2023generative,leinonen2020stochastic,harris2022generative,annau2023algorithmic}. These studies solely utilize high-resolution meteorological field data for supervision, learning the mapping process from low-resolution meteorological fields to high-resolution ones \cite{ren2023superbench}. However, the aforementioned methods are all direct applications of super-resolution models and do not incorporate special designs tailored to the characteristics of meteorological variables. They merely achieve downscaling results at specific magnifications following high-resolution supervisory data. Some recent works on simulating fluid field data \cite{fukami2023super} have attempted to achieve grid-independent continuous-resolution downscaling and have integrated physical information as prior \cite{esmaeilzadeh2020meshfreeflownet}. However, the relevant physical information and data are difficult to apply in the context of real-world data. Our recent work DeepPhysiNet \cite{li2024deepphysinet} bridges physical laws and deep learning for continuous weather modeling on real-world weather data. But above methods do not make use of sub-grid observational data as an auxiliary. Moreover, most studies focus solely on downscaling a single type of meteorological variables, such as temperature or precipitation, and are unable to simultaneously process multiple meteorological variables.

Addressing the issues present in previous downscaling efforts, we have specifically designed a continuous downscaling benchmark, combined with multi-scale observational data, tailored to the characteristics of meteorological variables. Our new model effectively recovers sub-grid states within coarse-resolution meteorological fields. Moreover, we have performed downscaling on five surface variables, which helps in obtaining more comprehensive meteorological state information at the scale of scatter stations.

\subsection{Image Super Resolution}
The widespread application of deep learning in meteorological downscaling is inseparable from the rapid development of image super-resolution tasks in the field of computer vision. The following is a brief introduction to the related work in image super-resolution (SR). 

The field of image super-resolution (SR) has witnessed a significant transformation with the advent of deep learning techniques \cite{wang2020deep}. Pioneering work, Super-Resolution Convolutional Neural Network (SRCNN) \cite{dong2015image}, demonstrated the effectiveness of deep learning for this task. Building on this foundation, many studies have proposed various super-resolution networks based on CNN-based encoder-decoder structures, enhancing the performance of super-resolution tasks \cite{kim2016accurate,lim2017enhanced,zhang2018image}. With the development of foundational models in vision, several super-resolution models based on GANs \cite{ledig2017photo,wang2018esrgan} and Transformers \cite{lu2022transformer, yang2020learning} have also been proposed. However, the aforementioned super-resolution models are only capable of achieving super-resolution at fixed magnifications. Inspired by implicit neural representations \cite{xie2022neural}, recent works have begun to explore super-resolution tasks with continuous resolutions \cite{chen2021learning,chen2022videoinr}. The aim is to achieve super-resolution at any arbitrary position by learning the mapping from coordinates to RGB values. However, super-resolution methods based on implicit neural representations suffer from a lack of sub-grid supervision, resulting in what is called continuous-resolution being merely a more sophisticated form of smoothing.

There is scarcely any existing work that has introduced super-resolution methods based on implicit neural representations into the realm of meteorological downscaling. Furthermore, unlike image super-resolution, meteorological variables often include a wealth of sub-grid station observational information, which can better assist models in learning information at continuous locations. Therefore, we incorporate scatter grid observations as auxiliary information into our benchmark, combined with high-resolution remote sensing observations, in the hope of effectively integrating multi-scale observational data to recover sub-grid meteorological states and achieve meteorological downscaling at scatter station scale.

\subsection{Hypernetworks}
Hypernetworks \cite{ha2016hypernetworks} are the type of model architecture that utilizes one network (commonly referred to as the hypernetwork) to predict the weights of another network (typically called the target network). Compared to traditional network architectures, hypernetworks offer more flexibility in their structure and input/output modalities. They have been widely applied across various fields such as computer vision, solving differential equations, and uncertainty quantification \cite{chauhan2023brief}. Leveraging the hypernetwork structure, the traditional per-sample optimization approach of implicit neural representations can be transformed into a data-conditioned hypernetwork learning architecture. This structure allows for the learning of the target network's parameters based on different input samples, and the design of the target network's input and output according to the requirements of the task. Regarding the application of hypernetworks to downscaling in meteorological fields, to the best of our knowledge, there are currently no similar efforts. Given the characteristics of hypernetwork structures, they are particularly suitable for meteorological data, which often comprises multi-modal data types. Therefore, the \emph{HyperDS} we propose utilizes the hypernetwork model structure and has been specifically designed to cater to the characteristics of the downscaling task.

\section{Problem Setting}
In this section, we will introduce the observation data-guided downscaling benchmark specifically designed based on our understanding of meteorological downscaling tasks. This includes a description of the benchmark, the datasets we used, and the evaluation metrics.

\begin{table*}[!htbp]
    \centering
    \caption{Meteorological variables used for downscaling.}
    \renewcommand\arraystretch{1.5}
    \resizebox{\linewidth}{!}{
    \begin{tabular}{c|c|c|c}
        \toprule
		{Long Name} &{Short Name} & {Description} & {Unit} \\
        \hline 
            10m u-component of wind & $u_{10}$ & Eastward component of the wind speed, at the height of 10 meters above the surface of the Earth. & $m/s$\\
        \hline
            10m v-component of wind & $v_{10}$ & Northward component of the wind speed, at the height of 10 meters above the surface of the Earth. & $m/s$\\
        \hline
            2m temperature & $t_{2m}$ &  Temperature of air at 2m above the surface of the land, sea or inland waters. & $K$\\
        \hline
            surface pressure & $sp$ &  Pressure of the atmosphere at the surface of land, sea and inland water. & $hPa$\\
        \hline
            total precipitation in 1 hour & $tp_{1h}$ &  Accumulated liquid and frozen water, comprising rain and snow, that falls to the Earth's surface in 1 hour. & $mm$\\
        \bottomrule
	\end{tabular}
 	}
	\label{tab:var}
\end{table*}

\subsection{Observation-Guided Weather Downscaling at Station-Scale}
\subsubsection{Task Description}
Unlike the objectives of previous downscaling or super-resolution tasks, which were to obtain high-resolution grid data, it is very important to capture the meteorological state at any given scatter station location, and this has significant practical value \cite{wu2023interpretable}. However, the results produced by most current meteorological tasks are structured gridded data, which need to be further processed through methods such as interpolation to obtain the state values at the scatter point locations of interest. As a result, simple interpolation without any learnable process creates an inherent bias between the gridded data and the scatter stations. Therefore, it is crucial to design specialized methods to effectively downscale gridded meteorological fields to scatter points and minimize the inherent bias between them.

To address this issue, inspired by data assimilation \cite{eyre2022assimilation,geer2021learning,cheng2023machine,chen2023towards}, we realize that gridded meteorological field data are obtained through the integration of multi-source, multi-scale observational data, and gridded model forecast result. Therefore, using observational data to guide the downscaling of meteorological fields is a very direct and reasonable approach. This allows us to recover sub-grid information of low-resolution gridded meteorological fields through multi-scale, multi-resolution observational data, thereby achieving the purpose of downscaling at scatter station scale.

Specifically, we classify the observational data into two categories: one is the gridded high-resolution indirect observational data (such as satellite observations), and the other is the scattered sub-grid direct observational data (such as weather observation stations). This also conforms to the observation type settings used in the operational assimilation forecasting in actual meteorological services \cite{geer2021learning}. Given multiple low spatial resolution meteorological fields data $\mathcal{F}_{input}\in \mathbb{R}^{1\times V\times LH\times LH}$ at a certain time step, gridded high-resolution indirect observational data sequence $\mathcal{O}\in \mathbb{R}^{T\times C\times TH\times TW}$ and scattered sub-grid direct observational data $\mathcal{S}\in \mathbb{R}^{1\times V\times\ N}$, where $V$ is the number of meteorological variables, $T$ and $C$ is the number of gridded observation frames and channels, $N$ is the number of scatter observations. our goal is to obtain the meteorological state values $\mathcal{F}_{output}\in \mathbb{R}^{1\times V\times\ M}$ at any $M$ scatter point locations by:
\begin{equation}
    \mathcal{F}_{output} = \Phi(\mathcal{F}_{input} | \Theta(\mathcal{O}),\mathcal{S})
\end{equation}
where $\Theta(\cdot)$ is a function that maps the indirect observational data into the meteorological variable domain. $\Phi(\cdot)$ is a downscaling model that is used for downscaling to the scatter station scale.  It should be noted that, in order to verify the generalization ability of the downscaling process for different scatter point locations, we require that the $N$ points in $\mathcal{S}$ and $M$ points in $\mathcal{F}_{output}$ are disjoint. Under such a setting, downscaling to multiple grid scales or random scatter point scales can be achieved by altering the positions of the target points.

\begin{table*}[!htbp]
    \centering
    \caption{Dataset used in this benchmark, period: 2017-01-01 to 2021-08-31.}
    \renewcommand\arraystretch{1.5}
    \resizebox{\linewidth}{!}{
    \begin{tabular}{c|c|c|c}
        \toprule
		{Data Name} &{Data Type} & {Resolution} & {Descriptions} \\
        \hline 
            ERA-5 Reanalysis \cite{hersbach2020era5} & Meteorological Field & $0.25^{\circ}$, 1 hour & Meteorological fields of 5 surface variables in Tab. \ref{tab:var}. \\
        \hline
            Himawari-8 L1 Gridded data \cite{bessho2016introduction} & High-resolution Gridded Observations & $5km$, 10 min & Rreflectance of channel 'albedo\_03', 'albedo\_05', 'tbb\_08' and 'tbb\_15'  \\
        \hline
            Weather2K \cite{zhu2023weather2k} & Scatter station Observations & sub-grid, 1 hour  & Observation state of air pressure, temperature, wind speed, and total precipitation in 1 hour.   \\
        \bottomrule
	\end{tabular}
 	}
	\label{tab:dataset}
\end{table*}

\subsubsection{Meteorological Variables Selection}
In order to select significant meteorological variables to verify the effectiveness of the downscaling method, we analyzed the characteristics of the observational data and specially selected five surface variables as the focus of our benchmark: $u$-component wind ($u_{10}$), $v$-component wind ($v_{10}$), 2-meters temperature $t_{2m}$, surface pressure ($sp$) and total precipitation in 1 hour ($tp_{1h}$). For details please refer to Tab. \ref{tab:var}. We chose these five variables primarily because the observational data includes the aforementioned variables, or the values of related variables can be roughly inferred through indirect observations, or there is an implicit correlation between the observational information and the variables.

\subsection{Dataset}
\begin{figure}
\centering
\includegraphics[width=\linewidth]{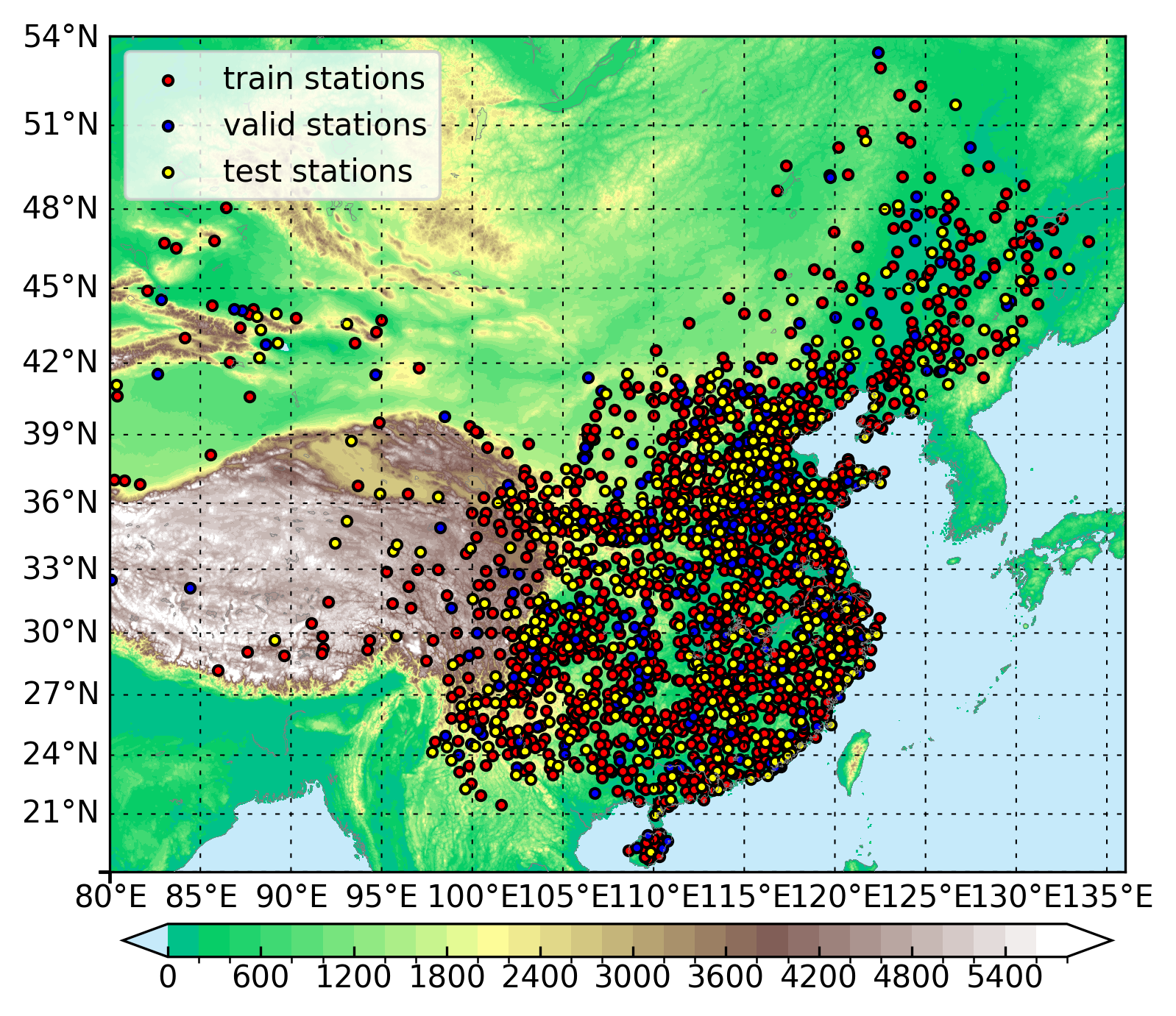}
\caption{The study area and scatter stations used in our paper. The red dots represent the training stations, the blue dots represent the validation stations and the yellow dots represent the test stations.}
\label{fig:study_area}
\end{figure}

This subsection will introduce the actual data used for the proposed benchmark, with the main data information available in Table \ref{tab:dataset}. Based on the observational data and meteorological field data we used, we selected the research area with a boundary of $80^{\circ}$E to $136^{\circ}$E and $18^{\circ}$N to $54^{\circ}$N, as shown in Fig. \ref{fig:study_area}. This area is an intersection of all the regions covered by all kinds of data we used.

\subsubsection{Meteorological Field Data}
We select the widely recognized ERA5 reanalysis data \cite{hersbach2020era5} as the meteorological field data. Its original resolution is $0.25^{\circ}$, covering the globe, with a temporal resolution of 1 hour. The ERA5 reanalysis data has been widely used in the field of meteorological forecasting based on deep learning \cite{rasp2020weatherbench,rasp2023weatherbench}, and it is employed as both the initial field and supervision data for models \cite{bi2023accurate,lam2023learning,chen2023fengwu,chen2023fuxi}. For our task, we have extracted data for the study area and, through the operation of average pooling, downsampled the data to a spatial resolution of $1^{\circ}$ to serve as input for the model. This setup enables us to provide high-resolution grid supervision.

It should be noted that previous downscaling work was mostly based on forecasting tasks, which primarily involved downscaling coarse-resolution forecast fields and utilizing high-resolution analysis data for supervision \cite{chen2023foundation,zhonginvestigating}. This setting is typically taken from the perspective of practical operational applications. Different from its starting point, our benchmark aims to study more effective downscaling methods at the scale of scattered stations based on deep learning. Hence, we hope to utilize readily available public data to provide as many complete samples as possible for model training. However, most model forecast fields and high-resolution analysis data are often difficult to obtain, or they have a low temporal resolution \cite{swinbank2016tigge}, making it challenging to meet the requirements for a large sample size. Therefore, we have chosen the most commonly used ERA5 reanalysis data, which ensures the fulfillment of our task requirements. The methods developed on it can also be well extended to situations where the forecast fields are used as inputs.

\subsubsection{Observation Data}
Various observation data are crucial to forming a structured grid of meteorological fields. In the field of meteorology, data assimilation tasks \cite{eyre2022assimilation,geer2021learning,cheng2023machine,chen2023towards} specifically study how to integrate different observation data into forecast fields, improving the performance of forecast models based on real-time observations. Integrating observational data effectively into meteorological fields to obtain more accurate meteorological states at different locations is an important research direction. We hope to enhance the accuracy of downscaling by incorporating observational information into downscaling tasks. On the one hand, using observational information to improve the accuracy of downscaling, and on the other hand, treating downscaling as a fundamental task to explore effective methods for integrating observational data into meteorological fields. 

Based on the task description provided earlier, we selected the L1 gridded data from the next-generation geostationary meteorological satellite Himawari-8 \cite{bessho2016introduction} as the gridded high-resolution indirect observational data, and we chose the station data set provided by the Weather2K dataset \cite{zhu2023weather2k} as the scattered sub-grid direct observational data. Below is a brief introduction to the two types of observational data:

\noindent \textbf{Himiwari-8 L1 Gridded Data} \cite{bessho2016introduction} is generated by JAXA/EORC from the Himawari Standard Data with re-sampling to equal latitude-longitude grids. It includes 16 spectral bands, with a spatial resolution of up to 2 km and a temporal resolution of 10 minutes, obtained from the Advanced Himawari Imager (AHI) onboard the Himawari-8 satellite. Due to storage constraints, we primarily downloaded the version with a 5 km spatial resolution and cropped the data to the research area of our interest. Furthermore, to reduce the memory footprint of the input data, we empirically selected four representative bands from the visible, near-infrared, and far-infrared spectral ranges. It should be noted that, as we are using Level 1 observational data rather than the results of the satellite retrieval of meteorological variables, the designed model is required to learn the state values of meteorological variables from the indirect satellite radiance values. This also implies that the retrieval process is inherently included in our task.

\noindent \textbf{Weather2K} dataset \cite{zhu2023weather2k} was originally designed for mesoscale weather forecasting tasks. It comprises hourly observations of 20 meteorological variables from 1866 ground observation stations across China, spanning from January 2017 to August 2021. We have selected four surface meteorological variables as the focus of our research, which are: air temperature, air pressure, wind speed, and precipitation in 1 hour. For the downscaling task at the scatter station scale, the primary objective is to verify the model's generalization performance at various scattered locations. Therefore, as shown in Fig. \ref{fig:study_area}, we randomly split the 1866 stations into three non-overlapping parts, with 1266 stations used for training, 200 stations for validation, and the remaining 400 stations used for testing.

\subsection{Evaluation Metrics}
To evaluate the downscaling effects of different methods at the scale of scattered observation stations, referencing prior work \cite{wu2023interpretable}, we have chosen the mean squared error (MSE) and mean absolute error (MAE) averaged over both stations and time as our assessment metrics. The calculation methods are as follows:
\begin{equation}
\begin{split}
    MSE &= \frac{1}{M*T}\sum_{i=1}^{M}\sum_{t=1}^T(Y_i^t - \hat{Y}_i^t)^2 \\
    MAE &= \frac{1}{M*T}\sum_{i=1}^{M}\sum_{t=1}^T|Y_i^t - \hat{Y}_i^t|
\end{split}
\end{equation}
where $M$ is the total number of test observations, $Y_i^t$ is the ground truth value for a given variable state at $i$-th station and $t$-th time point, $\hat{Y}_i^t$ is the predicted downscaled value. Both metrics are commonly used in regression tasks to measure the accuracy of the predicted values. Lower values of MSE and MAE indicate better model performance, with the MAE being particularly useful for understanding the error magnitude on an average per-observation basis.

\section{New Method}
In response to the downscaling task described above, we developed a novel method, namely \emph{HyperDS}, that effectively integrates high-resolution Himawari-8 satellite observations and scattered station observations to recover subgrid-scale meteorological states from low-resolution atmospheric fields, achieving continuous-resolution modeling of the meteorological field. This section will provide a detailed introduction to the structure and training strategy of our proposed method.

\begin{figure*}
    \centering
    \includegraphics[width=0.85\linewidth]{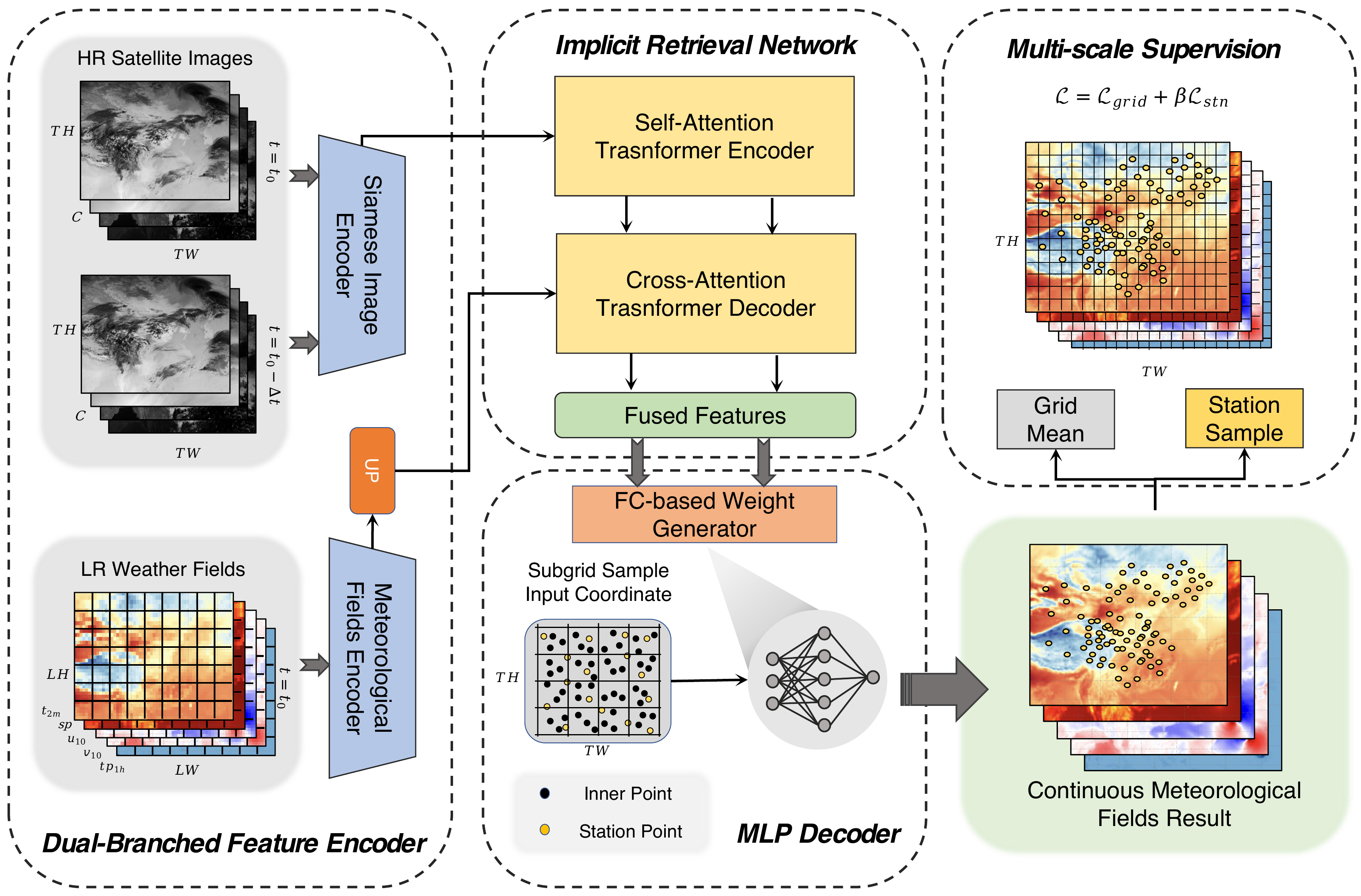}
    \caption{The proposed \emph{HyperDS} architecture. It mainly consists of three parts: a dual-branch feature encoder is used to extract semantic features from the input low-resolution meteorological field and high-resolution remote sensing images respectively; subsequently, the implicit retrieval network utilizes a cross-attention mechanism to implicitly fuse different feature information and align the remote sensing image features with meteorological field variables; and finally, the FC (fully connected)-based weight generator predicts the weight vector for the target network. The  MLP (Multi-Layer Perceptron) decoder, the target network, learns the mapping from the sampled subgrid coordinates to the corresponding location state values. It is supervised at both the observation station scale and the high-resolution grid scale, allowing for the continuous modeling of the meteorological fields.}
    \label{fig:model}
\end{figure*}

\subsection{Overall Structure}
The overall structure of \emph{HyperDS} can be viewed as a data-conditional hypernetwork architecture \cite{chauhan2023brief}. Considering the type of observational data, we use the indirectly observed high-resolution Himawari-8 satellite images as auxiliary input to the model, and the direct scattered observation station data as the model's station-scale supervision. The reason for this setup is to enable the model to learn the implicit meteorological field information from indirect remote sensing observations through operations such as encoding and feature extraction of the former. At the same time, supervision from the observation stations is used to correct the inherent biases that occur when downscaling from grid scale to station scale.

As illustrated in Fig. \ref{fig:model}, our model is composed of three sub-network structures: a dual-branch feature encoder based on CNN and an implicit inversion network based on a Transformer with cross-attention form the hypernetwork, which generates the fused features used to determine the weights of the target network; the target network, in turn, comprises an MLP-based decoder that learns the mapping from specific coordinates to meteorological states, using the weight parameters generated by the hypernetwork.

Additionally, in the input portion of the MLP decoder, we design a coordinate selection method based on subgrid sampling that more naturally and reasonably adapts to the capability of implicit neural representation for continuous state modeling. In such a case, by averaging the subgrid samples pixel by pixel, we can utilize high-resolution grid scale data for supervision and learn the deviation loss between the predicted values and the site-scale observations by sampling specific scatter station locations. Averaging and sampling of this form are also better aligned with the processing methods used to integrate multiscale observational data in meteorological fields.

\subsection{Dual-Branched Feature Encoder}
For processing meteorological field data and satellite imagery simultaneously, we implemented a simple dual-branch encoding structure based on ResNet-18 \cite{he2016deep} as the backbone for both the meteorological field encoder and the satellite image encoder. For the satellite image encoding branch specifically, we chose to input two adjacent Himawari-8 images in a siamese configuration into the encoder. The rationale behind this approach is inspired by previous work that utilized multi-frame images to infer wind fields by tracking cloud movements \cite{kazuki2017introduction}, with the hope that the model will autonomously learn the features of the two frames and some implicit gradient information. To balance the semantic information and spatial information of the features, we selected the features from the intermediate layers of the two encoders as the outputs of the model. Specifically: given the low-resolution meteorological $\mathcal{F}_{input}\in \mathbb{R}^{1\times 5\times LH\times LH}$ and two frames high-resolution Himawari-8 remote sensing images $\mathcal{O}\in \mathbb{R}^{2\times 4\times TH\times TW}$, the extracted features from each encoder can be computed by:
\begin{equation}
\begin{split}
    &F_{field} = {\rm Conv2d_{field}}({\rm MeteoEncoder}(\mathcal{F}_{input})) \\
    &F_{field} = {\rm Conv2d_{field}}(F_{field}) \\
    &F_{h8} = {\rm Concat}({\rm ImgEncoder}(\mathcal{O}_{0}), {\rm ImgEncoder}(\mathcal{O}_{1}) \\
    &F_{h8} = {\rm Conv2d_{field}}(F_{h8})
\end{split}
\end{equation}
The two features are extracted from the same stage in ResNet-18 and aligned across all dimensions into $C\times h\times w$ through upsampling and convolution operations.

\subsection{Implicit Retrieval Network}
To further integrate the two types of extracted features, we adopted a Transformer encoder-decoder network based on a cross-attention mechanism. This network implicitly retrieves indirect observation information into the meteorological field domain and effectively integrates it. We first flatten $F_{field}$ and $F_{h8}$ into shape $C\times hw$, and add the learnable positional encoding vector. Then, the features from high-resolution remote sensing images are fed to the self-attention Transformer encoder to further learn the relationships between different tokens. Following a cross-attention Transformer decoder receives features from both remote sensing images and meteorological fields to implicitly learn the relationships between different feature domains. Therefore, the fused features generated by the implicit retrieval network can be computed by:
\begin{equation}
    \begin{split}
        &F_{h8} = {\rm selfAttnEncoder}(F_{h8})\\
        &F_{fused} = {\rm crossAttnDecoder}(F_{h8}, F_{field})
    \end{split}
\end{equation}
Through the above calculations, the generated fusion features contain high semantic features of both low-resolution meteorological fields and high-resolution satellite observations, laying the foundation for subsequent continuous-resolution modeling.

\subsection{MLP Decoder with Subgrid-sampling}
\begin{figure*}
    \centering
    \includegraphics[width=0.95\linewidth]{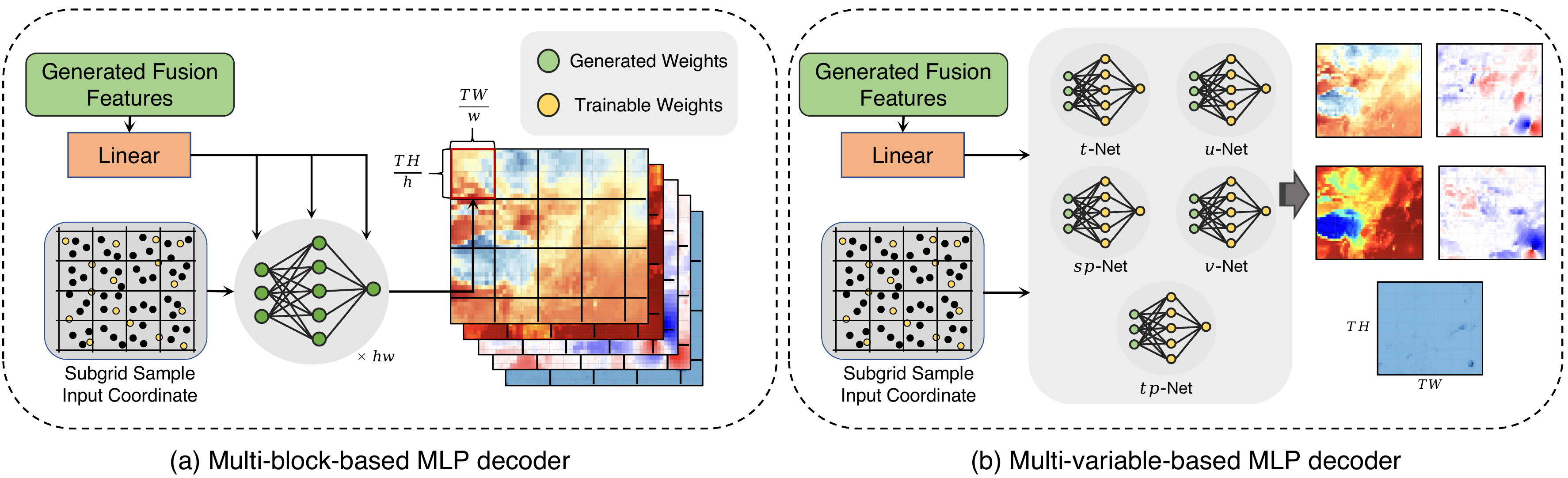}
    \caption{Two variants of MLP decoders based on implicit neural representations with subgrid sampling.}
    \label{fig:mlp_dec}
\end{figure*}

Based on the fusion features generated by the super network structure and combined with the latent neural representation method \cite{xie2022neural}, we designed a decoder module based on the Multilayer Perceptron (MLP). By learning the mapping from coordinate positions to meteorological states, we realized continuous modeling of the meteorological field. 

Unlike previous modeling methods based on latent neural representations that directly use the coordinates of the grid center to represent grid values \cite{chen2021learning, esmaeilzadeh2020meshfreeflownet}, we designed a specialized sub-grid sampling method specifically for meteorological fields. This allows us to construct a continuous representation of the meteorological field more naturally and reasonably. Specifically, for a given pixel $p$ in a high-resolution grid ${\rm Grid(TH, TW)}$, we randomly sample multiple inner points $I_p=\{(x_i,y_i)\ |\ i=1,2,\cdots, P\}$ within this pixel as the coordinate values to be input in this pixel. Then, we can obtain the state value for the corresponding resolution grid by calculating the average of the meteorological state values associated with all the inner points under that pixel, which can then be supervised using the corresponding high-resolution grid labels. Specifically, when a pixel includes the location of observation stations, we will also sample the positions of these observation stations. By using proposed the subgrid sampling method described above, we can average the data from observation stations into the mean value of the grid, thereby mitigating the conflict between the scattered point stations and the grid values. It should be noted additionally that, apart from the coordinate values, we also input the grid interpolation results of the state values corresponding to the coordinate points as auxiliary information into the MLP. For the sake of simplifying the expression, this part of the information will not be explicitly reflected in the following text.

As for the MLP modeling method used to represent the meteorological field, as shown in Fig. \ref{fig:mlp_dec}, we referenced previous work \cite{shaham2021spatially} and designed two different modeling structures. The first type, the multi-block-based MLP decoder, divides the target meteorological field into several blocks, each of which is continuously modeled by a separate MLP that takes input coordinates and simultaneously outputs the state values of all target variables. The weights of the network are obtained entirely through the linear mapping of fused features. Different from the first one, the second type is a multivariate MLP decoder that uses different MLPs to model the entire meteorological field of specific variables separately. The shallow parameters of each MLP are obtained through linear mapping of the fused features, whereas the deeper parameters are randomly initialized and constitute learnable weights. The reason for this setup is that the former, multi-block-based modeling approach, although more conducive to modeling high-frequency information, is more memory-intensive and increases model complexity with the number of sub-blocks. The latter has a relatively fixed computational complexity and memory usage, but it makes the modeling task more challenging for a single MLP. Therefore, each method has its advantages and disadvantages, and we hope to provide more flexible options for our approach.

\subsection{Loss Function}
The supervision label data proposed for \emph{HyperDS} includes two scales: grid-scale and station-scale. By inputting different sampling coordinates into the MLP decoder, predictions for the corresponding scales can be generated. Detailed introductions follow.

\subsubsection{Grid-scale Loss} 
Referencing previous downscaling work based on Super-Resolution (SR), we use high-resolution meteorological fields as supervision at the grid scale. However, unlike previous work, we obtain the prediction result for a target pixel by calculating the average of the sub-grid inner points within that pixel. To be specific, given the high-resolution grid label field $\mathcal{F}_{label}^{grid}\in \mathbb{R}^{1\times 5\times TH\times TW}$, we sample $P$ inner points $I_p=\{(x_i,y_i)\ |\ i=1,2,\cdots, P\}$ at each pixel $p$ in ${\rm Grid(TH, TW)}$ and the predicted field can be computed by:
\begin{equation}
    \mathcal{F}_{output}^{grid} = \frac{1}{P}\sum_{i=1}^P{\rm mlpDecoder}(I, F_{fused})
\end{equation}
then, the grid-scale loss can be computed by:
\begin{equation}
    \mathcal{L}_{grid} = \Vert \mathcal{F}_{label}^{grid} - \mathcal{F}_{output}^{grid} \Vert ^2
\end{equation}
However, while high-resolution supervision can provide accurate mean supervision on fine grids, in actual applications, high-resolution grid supervision is often difficult to obtain. To account for station-scale downscaling in such situations, we referenced previous work on modeling dynamic systems \cite{rao2023encoding} and designed a grid loss function for when high-resolution gridded supervision is not available. In such cases, still benefiting from the sub-grid sampling coordinate input form, we could also compute the mean values covered by the low-resolution pixel in ${\rm Grid(LH, LW)}$, at the same time interpolate the input low-resolution fields into a fine-grained one. Thereby simultaneously obtaining the interpolated high-resolution supervision as well as the low-resolution mean supervision:
\begin{equation}
    \begin{split}
        &\mathcal{L}_{HR} = \Vert {\rm Interp}(\mathcal{F}_{input}) - \mathcal{F}_{output}^{grid} \Vert ^2\\\label{con:wo_gt}
        &\mathcal{L}_{LR} = \Vert \mathcal{F}_{input} - {\rm Avgpool}(\mathcal{F}_{output}^{grid}) \Vert ^2\\
        &\mathcal{L}_{grid} = \mathcal{L}_{HR} + \mathcal{L}_{LR}
    \end{split}
\end{equation}
Through such an approach, in the absence of high-resolution grid supervision information, it is possible to provide as much grid supervision information as possible to the maximum extent.

\subsubsection{Station-scale Loss}
To integrate the observational information from scattered stations into the downscaling process and mitigate the inherent bias between the meteorological field and grid observations, the training process incorporates station-scale supervision to learn the cross-scale mapping from the meteorological field to the stations. To be specific, given the station-scale label $\mathcal{F}_{label}^{station}\in \mathbb{R}^{1\times 5\times M}$, where $M$ is the number of stations, we could sample the corresponding coordinates and computed the station scale output by:
\begin{equation}
    \mathcal{F}_{output}^{station} = {\rm mlpDecoder}(I', F_{fused})
\end{equation}
and the station-scale loss is
\begin{equation}
    \mathcal{L}_{stn} = \Vert \mathcal{F}_{label}^{station} - \mathcal{F}_{output}^{station} \Vert ^2
\end{equation}
It should be noted that since only the wind speed variable, rather than its components, is provided in the Weather2K data \cite{zhu2023weather2k}, we base our calculation of wind speed loss on the following formula:
\begin{equation}
    wind\_speed = \sqrt{u_{10}^2 + v_{10}^2}.
\end{equation}
Combining the above loss functions, we can derive the final loss function as:
\begin{equation}
    \mathcal{L} = \mathcal{L}_{grid} + \beta\mathcal{L}_{stn} \label{con:loss_func}
\end{equation}
where $\beta$ are the loss coefficients that need to be manually set.

\section{Results}
In this section, we design experiments and baseline methods tailored to the downscaling task we have constructed, which will be compared with our proposed \emph{HyperDS} model.

\subsection{Experiment Details}
To validate the downscaling performance from low-resolution meteorological fields to arbitrary scatter stations, we downsampled the original ERA5 reanalysis data \cite{hersbach2020era5} to a spatial resolution of $1^{\circ}$ using the method of average pooling, which served as the input meteorological field data. Moreover, we used the original ERA5 data at $0.25^{\circ}$ resolution as high-resolution grid supervision. Additionally, we conducted experiments without high-resolution grid supervision by setting the loss according to Eq. \ref{con:wo_gt}. The dataset was divided into three parts in chronological order: data from January 1, 2017, to August 31, 2020, was used for training; data from September 1, 2020, to December 31, 2020, was used for validation; and data from January 1, 2021, to August 31, 2021, was utilized for testing. It should be noted that for the observational station data from Weather2K \cite{zhu2023weather2k}, we partitioned the dataset based on both time and station locations. This division requires that the model exhibit strong generalization performance both temporally and spatially when validated at the observational station scale, which presents a significant challenge. Therefore, the experiments we set up aim to recover the meteorological state variables at a scatter scale from a $1^{\circ}$ spatial resolution meteorological field and to validate the performance at 400 randomly sampled test stations (as shown in Fig. \ref{fig:study_area}) within the testing period. The hyperparameter number of samples in the MLP decoder is set to 10, and the loss coefficient $\beta=0.05$.

To ensure a fair comparison of different methods, we have established the same training procedures and hyperparameters for all. We optimized the model using the Adam optimization method \cite{kingma2014adam}, employing a cyclical learning rate with cosine annealing \cite{loshchilov2016sgdr}, starting with an initial learning rate of 0.0001, for a total of 50 epochs of training. We choose the checkpoint with the lowest station-level loss in the validation sets for testing. We trained our proposed model using 4x NVIDIA A100 GPUs, setting the batch size to 4 per GPU.

\subsection{Baselines}
For the benchmark we proposed, we designed two basic baseline methods for comparison to validate the effectiveness of our method. The following subsections provide a detailed introduction to these methods.
\subsubsection{Interpolation of Weather Field into Station Scale}
One of the simplest and most direct methods to obtain meteorological state variables at the scale of scattered stations from a meteorological field is through interpolation. We use the ${\rm DataArray.interp}$ function with the default setting from the open-source ${\rm xarray}$ library to perform interpolation on the meteorological field, based directly on the absolute positions of latitude and longitude, with each grid cell's state value corresponding to its center point coordinates. We mainly performed interpolation on meteorological fields with resolutions of $1^{\circ}$ and $0.25^{\circ}$, corresponding to the interpolated results from the input low-resolution meteorological fields and the high-resolution supervision. Since the interpolation process does not incorporate any available observational information, this method can serve as our most basic baseline result. Moreover, it can reflect to some extent the inherent bias that exists between the meteorological fields and observations.

\subsubsection{Super-Resolution-based Downscaling with Observations}
Given the widespread application of super-resolution models in downscaling tasks, we specifically modified traditional super-resolution models for our proposed benchmark, integrating multi-scale observational information into them. As shown in Fig. \ref{fig:overview}(a), traditional SR-based downscaling methods mainly learn the mapping between low-resolution input fields and the target high-resolution fields within an encoder-decoder architecture. We adopted a straightforward approach to incorporate high-resolution Himawari-8 (H8) satellite imagery observations and Weather2K station observations into the model. Specifically, given the high-resolution H8 images we encode them with a single convolutional layer, then align the dimension with the input meteorological fields by average pooling operation and concatenated them on top of the input low-resolution meteorological field before feeding them into the super-resolution model. After obtaining the meteorological field at the target resolution, the meteorological state of the scattered stations can be acquired through interpolation. Subsequently, both the grid supervision and the corresponding station supervision data are utilized to compute the loss function, which is then used for backpropagation. In terms of the choice of super-resolution models, we selected the classic UNet-based super-resolution model \cite{ronneberger2015u} and the EDSR \cite{lim2017enhanced} and made modifications to their structures and implementations. 


\subsection{Comparison with Baselines}
We compared the performance of our proposed \emph{HyperDS} method with other baseline methods for station-scale downscaling with high-resolution grid supervision. Due to the trade-off between grid-scale supervision and station-scale supervision in the model optimization process, but as our current task mainly focuses on the performance at the station scale, we chose the checkpoint with the smallest station-scale loss on the validation set during the training process as the model for testing. The loss coefficients can also greatly affect site performance, so considering the need to balance the losses at both scales, we empirically set the loss coefficients in Eq. \ref{con:loss_func} as $\beta=0.05$. Subsequent sections will further discuss the trade-off problem between the losses at the two scales. Since the Weather2K dataset \cite{zhu2023weather2k} used as labels do not include the wind speed component, we compared the downscaling results of 4 surface variables, which are: wind speed ($ws$), surface pressure ($sp$), 2m temperature ($t_{2m}$) and total precipitation in 1 hour ($tp_{1h}$). 

\begin{table*}[!htbp]
    \centering
    \caption{Station-level downscaling results for wind speed ($ws$), surface pressure ($sp$), 2m temperature ($t_{2m}$) and total precipitation in 1 hour ($tp_{1h}$) of various methods.}
    \renewcommand\arraystretch{1.5}
    \resizebox{0.9\linewidth}{!}{
    \begin{tabular}{c|cccccccc}
    \toprule
		 \multirow{2}{*}{Method} & \multicolumn{2}{c}{$ws$} & \multicolumn{2}{c}{$sp$} & \multicolumn{2}{c}{$t_{2m}$} & \multicolumn{2}{c}{$tp_{1h}$} \\
   
         & MSE & MAE & MSE & MAE & MSE & MAE & MSE & MAE\\
    \hline
        ERA5\ $1^\circ$ & 5.5642 & 1.7842 & 1048.3886 & 21.1125 & 7.7483 & 1.9235 & 1.1396 & 0.1896 \\
   
        ERA5\ $0.25^\circ$  & 6.2164 & 1.9118 & 801.3915 & 17.2155 & 6.7855 & 1.7770 & 1.2018 & 0.1893\\
    
        UNet \cite{ronneberger2015u}& 5.4575 & 1.7757 & 967.8221 & 20.1538 & 7.3537 & 1.8806 & 1.1426 & 0.1955  \\

        EDSR \cite{lim2017enhanced} & 6.1547 & 1.8905& 896.9313 &19.2898 & 7.1336 & 1.8386 & 1.1572 & 0.1982 \\
    \hline
        HyperDS (Ours, multi-var) & \textbf{1.7995} & \textbf{0.9568} & 716.0126 & 15.4001 & \textbf{6.3588} & \textbf{1.7656} & 1.1278 & 0.1887\\
        HyperDS (Ours, multi-block) & 1.9671 & 1.0126 & \textbf{645.0722} & \textbf{14.6524} & {6.5747} & 1.8400 & \textbf{1.1260} & \textbf{0.1859}\\
    \bottomrule
    \end{tabular}}
  
	\label{tab:stn_var_metric}
\end{table*}

\subsubsection{Test Results by Variables}
Tab. \ref{tab:stn_var_metric} displays the test metrics for different variables using different methods at the 400 testing observation stations. From the results, it can be seen that our proposed \emph{HyperDS} method outperforms the compared baseline methods on all variables. Particularly for wind speed and surface pressure, our method significantly exceeds the others, with the MSE for wind speed improving by 67\% and for surface pressure by 19.5\% compared to the best baseline results.

It should be noted that in the results of direct interpolation of meteorological fields, the ERA5 $1^\circ$ interpolation results are superior to the $0.25^\circ$ interpolation results in metrics such as wind speed and precipitation, which seems counterintuitive. However, similar results have been reported in recent related work \cite{wu2023interpretable}. We believe this is due to the limited assimilation of observational station data in the ECMWF reanalysis data for the China region.

Upon further analysis of the test results for different variables, it is evident that our method shows the most significant improvement in wind speed. This is primarily because wind speed exhibits the most notable sub-grid variability, with local wind speeds often being influenced by a variety of small-scale meteorological processes such as turbulence, making it difficult to capture at the relatively coarse resolution of grid scales. On the contrary, for variables such as 2m temperature and precipitation, the improvement from our method is relatively small. This is because the variability of local temperature is relatively gradual, and as for precipitation, due to its sparse and long-tailed distribution \cite{hess2022physically}, acceptable downscaling results can be obtained by simply interpolating the grid data.

For the two super-resolution-based comparison methods, we can see that although observational information has been incorporated, the improvement in overall station downscaling performance is quite limited. We believe that this is because traditional super-resolution methods, which are based on fixed-resolution grid supervision, place more emphasis on the regression task for each grid pixel, and the learning process remains a discrete mapping from coarse grid scales to fine grid scales. In contrast, the optimization goal of our proposed \emph{HyperDS} method, which is based on hypernetworks and implicit neural representations, is to learn a mapping from arbitrary coordinates to meteorological states, constructing a continuous representation of the meteorological field. This endows the model with stronger capabilities for expressing sub-grid-scale information. This also explains why the EDSR model, which has stronger grid super-resolution capabilities, performs worse in station downscaling than the simpler UNet model. The reason is that EDSR has a stronger ability to fuse and extract features at the grid scale, which makes it difficult for the model to generalize well to the station scale, even with the inclusion of station-scale observations as supervisory labels.

For the two different MLP decoder structures we proposed, it can be seen that different decoder outcomes have certain performance differences for different variables. The multivariate-based HyperDS performs relatively better on wind speed and 2m temperature, while the multi-block-based HyperDS performs better on surface pressure and precipitation. We believe this is due to the different statistical distributions of various variables. Although we have normalized all variables by their mean and variance, making them roughly follow a Gaussian distribution with zero mean, there are still significant differences in the value ranges of each variable after normalization. For example, the variation range of surface air pressure is relatively larger, whereas the temperature variation range is comparatively smaller. The decoder based on multiple blocks, with each MLP representing a local region, can model the data for a specific area more effectively, thus avoiding the issue of too great a range of variable changes caused by global modeling. The decoder based on multiple variables uses a single MLP to model the entire meteorological field of a region, which is more effective for variables with smaller ranges of variation and can also save more computational memory consumption.

\subsubsection{Result Visualization}
\begin{figure*}
    \centering
    \includegraphics[width=0.95\linewidth]{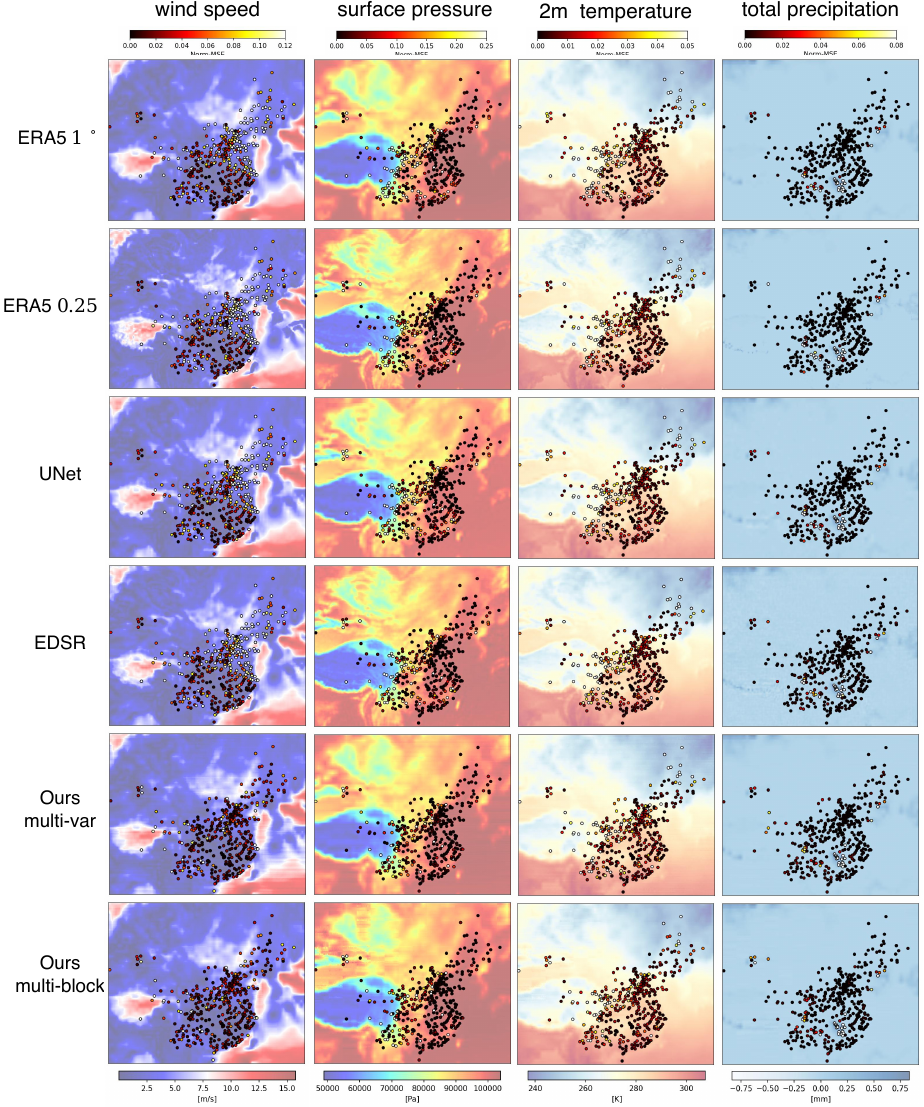}
    \caption{Visualization comparison of downscaling to station-scale using different methods, where the color of each station represents the magnitude of the normalized mean square error at that station, with lighter colors indicating larger errors, i.e. the darker the color of the site, the better the performance of the downscaling. The base map of each visual image is also the result of downscaling at the grid scale with $0.25^\circ$ spatial resolution of each model.}
    \label{fig:comp_vis}
\end{figure*}

Fig. \ref{fig:comp_vis} illustrates the results of downscaling at different test stations using various methods, where the color of each station represents the magnitude of the normalized mean square error at that station, with lighter colors indicating larger errors. The base map of each visual image is also the result of downscaling at the grid scale with $0.25^\circ$ spatial resolution of each model. The results in the figure show that our proposed \emph{HyperSR} method performs significantly better at the site scale compared to other methods. In particular, for the wind speed variable, the baseline methods exhibit a large error in the northeastern area of the study region (more white dots), but our method can effectively correct the downscaling bias in this area. Combining this with Fig. \ref{fig:study_area}, we can see that for areas with sparser training stations (such as the northeastern and western regions), the downscaling performance of all methods tends to decline to different extents compared to the densely observed southeastern region. However, our method still exhibits better generalization performance; for instance, for the 2m temperature variable, our method has relatively fewer white dots in the northeastern region compared to other baseline methods. It is noteworthy that the downscaling method based on super-resolution (SR) achieves better downscaling results at the grid scale (i.e., base map) compared to our method. This is a limitation of our method and a direction for further improvement in the future.

\subsection{Downscaling without High-resolution Gridded Supervision}
In practical applications, high-resolution gridded meteorological fields are often difficult to obtain, whereas direct observations at observation stations are relatively easier to access and the data quality is more stable. Obtaining station-scale meteorological states directly from low-resolution meteorological fields is a very meaningful task. Hence, we conducted station-scale downscaling experiments without high-resolution grid supervision based on the designed loss function Eq. \ref{con:wo_gt}.

Tab. \ref{tab:wo_hrs} shows the downscaling results at the station level for different methods. The results from the table indicate that, even without high-resolution grid supervision, our proposed HyperDS outperforms other methods on the majority of the evaluation metrics for most variables, particularly for the wind speed variable. However, for 2m-temperature and surface pressure variables, compared to the results supervised with high-resolution grid data, there is a noticeable performance decrease (for surface pressure, the MSE decreased from 645.0722 to 805.9112, and for 2m-temperature, it decreased from 6.3588 to 7.1487). This performance degradation is because high-resolution grid supervision provides a significant improvement over coarse-resolution grid inputs at the station level for these two variables. Consequently, the absence of high-resolution grid data supervision leads to a decline in performance. In contrast, for the wind speed variable, the downscaling performance at the station level is slightly improved (MSE decreased from 1.7995 to 1.7815) because the interpolation results from low-resolution inputs are less dependent on high-resolution supervision.

Although there is a certain degree of performance decline, our proposed HyperDS method still outperforms the comparison methods even without high-resolution grid supervision. Even when the comparison methods incorporate high-resolution supervision (as shown in the results of Tab. \ref{tab:stn_var_metric}), our method remains superior on most metrics compared to those based on grid super-resolution networks.

\begin{table*}[!htbp]
    \centering
    \caption{Station-level downscaling results for wind speed ($ws$), surface pressure ($sp$), 2m temperature ($t_{2m}$) and total precipitation in 1 hour ($tp_{1h}$) of various methods without high-resolution gridded supervision.}
    \renewcommand\arraystretch{1.5}
    \resizebox{0.85\linewidth}{!}{
    \begin{tabular}{c|cccccccc}
    \toprule
		 \multirow{2}{*}{Method} & \multicolumn{2}{c}{$ws$} & \multicolumn{2}{c}{$sp$} & \multicolumn{2}{c}{$t_{2m}$} & \multicolumn{2}{c}{$tp_{1h}$} \\
   
         & MSE & MAE & MSE & MAE & MSE & MAE & MSE & MAE\\
    \hline
    
        UNet \cite{ronneberger2015u}&5.3051&1.7508&1078.9179&21.6993&7.6797&1.9225&1.1426&0.1992  \\

        EDSR \cite{lim2017enhanced} &5.2994&1.7495&1082.5246&21.7704&7.6539&1.9183&1.1426&\textbf{0.1902} \\
    \hline
        HyperDS (Ours, multi-var) & \textbf{1.7815} & \textbf{0.9613} & 901.4435 & 17.9062 & \textbf{7.1487} & \textbf{1.8239} & 1.1319 & {0.1940} \\
        HyperDS (Ours, multi-block) & 2.0379 & 1.0335 & \textbf{805.9112} & \textbf{16.0210} & 7.4909 & 1.9281 & \textbf{1.1274} & {0.1943}\\
    \bottomrule
    \end{tabular}}
  
	\label{tab:wo_hrs}
\end{table*}

\subsection{Ablation Study}
We further compared the performance of the \emph{HyperDS} method under different settings to verify the impact of the inclusion of observational data and the number of samples on the method's performance.
\begin{table*}[!htbp]
    \centering
    \caption{Ablation study results of HyperDS with multi-block-based MLP decoder. The settings 'station' and 'h8' represent the station-level supervision and H8 satellite images input. The setting 'sample' represents the subgrid-sampling strategy in MLP Decoder.}
    \renewcommand\arraystretch{1.5}
    \resizebox{0.9\linewidth}{!}{
    \begin{tabular}{c|ccc|cccccccc}
    \toprule
		 \multirow{2}{*}{Method} & \multicolumn{3}{c|}{Settings}& \multicolumn{2}{c}{$ws$} & \multicolumn{2}{c}{$sp$} & \multicolumn{2}{c}{$t_{2m}$} & \multicolumn{2}{c}{$tp_{1h}$} \\
   
         &station&h8&sample& MSE & MAE & MSE & MAE & MSE & MAE & MSE & MAE\\
    \hline
        \multirow{4}{*}{HyperDS}&\ding{55}&\ding{51}&\ding{51}&5.4509&1.7769&1088.7840&23.3697&9.2645&2.2367&1.1427&0.1983\\
   
        & \ding{51}&\ding{55}&\ding{51}& \textbf{1.8876} & \textbf{0.9893} & 721.9860 & 15.2776 & 6.8517 & 1.8417 & 1.1262 & 0.1910\\
    
        & \ding{51}&\ding{51}&\ding{55}& 2.3029 & 1.1133 & 729.5865 & 15.7978 & 7.1172 & 1.8925 & 1.1322 & 0.1936  \\
        & \ding{51}&\ding{51}&\ding{51}& 1.9671  & 1.0126 & \textbf{645.0722} & \textbf{14.6524} & \textbf{6.5747} & \textbf{1.8400} & \textbf{1.1260} & \textbf{0.1859}  \\
    \bottomrule
    \end{tabular}}\label{tab:ablation_study}
\end{table*}

Tab. \ref{tab:ablation_study} shows the test performance of \emph{HyperDS} under different experimental settings. The results indicate that the inclusion of station observation supervision is the most critical factor affecting the model's performance. This is intuitive, as previous work has also shown that there is an inherent bias between the meteorology itself and scatter station observations \cite{wu2023interpretable}, which cannot be recovered solely through high-resolution grid supervision. Based on this result, coupled with the fact that we use station observations as our supervision labels, it means that no real-time station observations are needed during the model inference stage. The model itself can adaptively generalize the meteorological field to any station location, which also implies that the model has learned the inherent bias from the meteorological field to the station and has effectively reduced it. Therefore, our method can be regarded as a general interpolation model from the meteorological field to stations.

Regarding the input of H8 remote sensing satellite images, although it is not a direct representation of meteorological conditions, through our designed feature extraction and implicit retrieval network, the model can learn useful information from indirect observations. However, the results indicate that the input from H8 did not show a positive impact on all variables. This is because the satellite observations we input are Level 1 radiance data, in which the meteorological state information is implicit and incomplete. From the types of Level 2 (L2) inversion products provided by the Himawari-8 satellite, it is evident that the primary meteorological variables related to it are surface temperature, humidity, and high-altitude wind speed (obtained indirectly based on cloud movement). Therefore, in the experimental results, the incorporation of H8 data has a more pronounced improvement in 2m temperature and surface pressure (which are strongly correlated with humidity). However, for the surface wind speed, since it has a significant deviation from the high-altitude wind speed, the results do not show a direct enhancement.

Sampling at the subgrid coordinates is also one of the innovative aspects of our method. By using subgrid sampling, the traditional implicit neural representation methods can be better aligned with the continuous distribution characteristics of the meteorological field, and the scattered station observations are treated as samples to be averaged with other samples within the same pixel. We configured various subgrid sampling numbers for comparison, and the experimental results also indicate that more samples typically yield better experimental outcomes and faster convergence rates. However, excessive sampling usually means greater GPU memory usage; therefore, we only set the maximum number of samples to 10 during the experimental process.

\subsection{Optimization process analysis}
Since our method requires the simultaneous optimization of two losses at the grid scale and the station scale, we further discuss the trade-off between these two types of losses to analyze the impact of different losses and labels on the model optimization process.
\begin{figure}[htbp]
    \centering
    \includegraphics[width=0.95\linewidth]{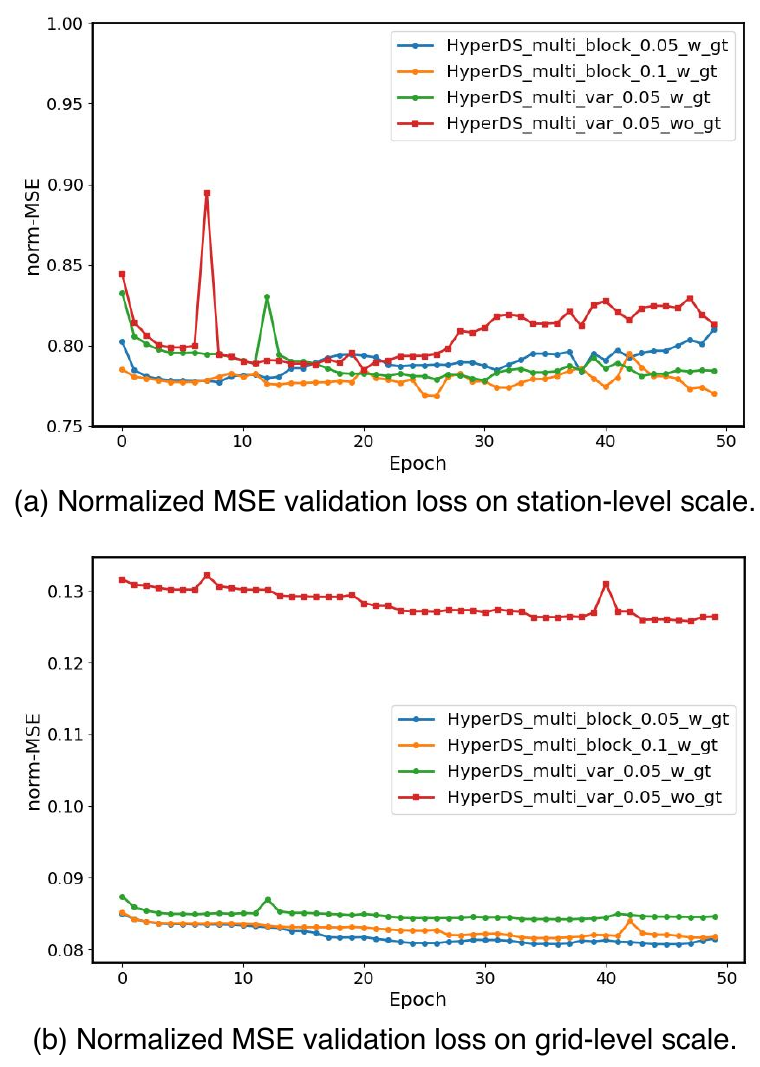}
    \caption{Illustration of the changes in the normalized MSE loss on the validation set under various hyperparameter settings during the training process, with 0.05 and 0.1 indicating the $\gamma$ values of the station loss in Eq. \ref{con:loss_func}.}
    \label{fig:trade_off}
\end{figure}
Both previous work \cite{wu2023interpretable} and the experimental results discussed above have adequately illustrated a substantial systematic bias between meteorological conditions at the station scale and the grid scale. Our approach uses the meteorological fields at the grid scale as input and, by modeling a continuous representation of the meteorological field, aims to obtain high-precision meteorological states at the station scale to alleviate this systematic bias. 
However, during the training process, as mentioned in Eq. \ref{con:loss_func}, we introduced supervisory information from both the station scale and the grid scale and combined them in a weighted sum to serve as the objective function. This also creates a trade-off between the two different losses. 

Fig. \ref{fig:trade_off} displays the changes in the normalized MSE loss for the validation set at both the station-scale and grid scale during the training process. It should be noted that because we have incorporated strong prior information (such as interpolation results, etc.) as inputs into our model, the model is capable of achieving satisfactory convergence within just one epoch. As a result, the overall loss function appears relatively smooth. The curves in the figure indicate that for all models, the grid-scale loss decreases steadily with the progression of training. However, the station-scale loss is relatively more volatile and tends to first decrease and then increase as the number of training epochs increases (this is particularly evident for the models with $\beta=0.05$). As $\beta$ increases from 0.05 to 0.1, this phenomenon is somewhat mitigated, but there is a corresponding decline in performance at the grid scale. We believe that such results are primarily due to a certain level of discrepancy between the two types of losses, and the fact that there are fewer station-scale samples compared to grid-scale samples, leading to a degree of sample imbalance. Furthermore, the results in the table also reveal that high-resolution grid supervision has a significant impact on grid-scale performance, but the effect is relatively tolerable for station-scale predictions. Upon further analysis of cases with high-resolution station supervision, models with multi-block decoders exhibit a significant performance improvement at the grid scale compared to multi-variable decoders; however, this improvement is not as pronounced at the station scale. Additionally, the convergence speed of the former is noticeably superior to that of the latter.

From the analysis above, it is evident that for the novel benchmark we proposed, it is unreasonable to focus solely on improving station-scale performance without considering the grid scale. Therefore, devising more rational model structures to further enhance the downscaling performance at both scales is an important research direction for the future.

\section{Discussion}
The purpose of this paper is to downscale grid-scale meteorological field data to the scale of discrete scatter stations. This allows us to obtain the meteorological state at any location based on widely used coarse-resolution meteorological field data, which has significant practical importance. The benchmark proposed in this paper, along with the novel method HyperDS, integrates multi-scale observational information into the downscaling task, effectively achieving station-scale downscaling. However, there are still many aspects of this task that can be further explored and researched. We will discuss the current issues and potential future directions from two perspectives: the data and the methodology.

\subsubsection{Observation Data}
Based on the experiments and analyses conducted, it is evident that incorporating multi-scale observational data plays a crucial role in the downscaling task. The dataset we currently propose includes only two types of observational information: station observations and geostationary satellite radiance values. Especially with regard to satellite observations, the data types contained within a single data source are quite limited. In the operational forecasting data assimilation process \cite{eyre2022assimilation}, multisource and multisensor remote sensing data are used to obtain different meteorological variables. Therefore, integrating more types of remote sensing observation data into the dataset is important for constructing a continuous-scale multivariate meteorological field. However, this task is also very challenging, involving specialized knowledge related to the sensors and corresponding meteorological variables, as well as complex data preprocessing procedures. In addition, the number of station data points we utilized is still relatively small. Previous work has used data from tens of thousands of stations to achieve high-accuracy station-scale weather forecasts \cite{wu2023interpretable}. Incorporating more station data is also crucial for continuous-scale modeling. We also welcome more researchers to integrate more types of data into our benchmark to enhance the capability of continuous-scale modeling for more meteorological variables, and we hope that related work can assist with the practical applications of meteorological forecasting.

\subsubsection{Models and Methodology}
Designing model architectures and methods suitable for the current task is also key to improving the performance of the benchmark. Unlike traditional downscaling works based on super-resolution, downscaling to the scale of individual stations places higher demands on the model's ability to represent resolution continuously. The \emph{HyperDS} proposed in this paper starts from this perspective and has achieved good performance in downscaling to the station scale. However, the current method sacrifices the performance of grid-scale downscaling to some extent in order to enhance station-scale downscaling, which is not a trade-off we desire to see. This is also a common problem with many super-resolution methods based on implicit neural representations. Therefore, how to design a more effective model that ensures multi-scale modeling accuracy while supporting continuous-resolution representation is an urgent issue to be addressed in the future.

\section{Conclusion}
In this paper, we extend the traditional fixed-resolution grid-based downscaling task to the scale of scattered station scale, based on the characteristics of meteorological variables. Inspired by data assimilation \cite{eyre2022assimilation}, we integrate multi-scale observational data into the downscaling process and build a novel benchmark and dataset that downscales coarse-resolution meteorological fields to station scales. Building on this foundation, we propose a new model based on a hypernetwork structure called \emph{HyperDS}. It uses high-resolution remote sensing images as prior input and scattered observation station data as station-scale labels. By continuously modeling the meteorological field, it effectively integrates multi-scale observational information and achieves high-precision meteorological field downscaling at the station scale. Through extensive experimental comparisons with specially designed baseline methods, we have verified the effectiveness of our proposed approach, particularly in terms of performance on wind speed and surface pressure variables, where it significantly outperforms other methods. This paper represents the first exploration into observation-driven downscaling of meteorological fields to station scales. We hope that in the future, more researchers will build on this foundation to study more effective methods, enhancing the accuracy and capability of continuous meteorological field modeling.

 
%
\bibliography{ref}
\bibliographystyle{IEEEtran}

\vfill

\end{document}